\documentclass[runningheads]{llncs}
\usepackage[T1]{fontenc}
\usepackage{graphicx,verbatim}

\usepackage{amsmath}
\usepackage{booktabs}  
\usepackage{multirow}  
\usepackage{amssymb}  
\usepackage{graphicx}  
\usepackage{color}
\usepackage{hyperref}
\usepackage{cite}

\begin{document}

\title{A Unified Framework for Joint Detection of Lacunes and Enlarged Perivascular Spaces}

\author{Lucas He\inst{1,2}\thanks{Corresponding author.} \and
Krinos Li\inst{3} \and
Hanyuan Zhang\inst{1} \and
Runlong He\inst{1} \and
Silvia Ingala\inst{4} \and
Luigi Lorenzini\inst{5} \and
Marleen de Bruijne\inst{6} \and
Frederik Barkhof\inst{1,5,7} \and
Rhodri Davies\inst{2,8,9}\textsuperscript{,$\dagger$} \and
Carole Sudre\inst{1,2,8}\textsuperscript{,$\dagger$}}

\authorrunning{L. He et al.}

\institute{Hawkes Institute, University College London, UK \and
Unit for Lifelong Health and Aging, University College London, UK \and
Bioengineering Department and Imperial-X, Imperial College London, UK \and
Department of Diagnostic Radiology, Copenhagen University Hospital, Denmark \and
Department of Radiology \& Nuclear Medicine, Amsterdam UMC, Vrije Universiteit, The Netherlands \and
Department of Radiology and Nuclear Medicine, Erasmus MC, Rotterdam, The Netherlands \and
Queen Square Institute of Neurology, University College London, UK \and
Institute of Cardiovascular Sciences, University College London, UK \and 
Barts Heart Centre, St Bartholomew's Hospital, London, UK \\
\email{Lucas.he.23@ucl.ac.uk}}

\maketitle
{\let\thefootnote\relax\footnotetext{$^{\dagger}$ Joint senior authors.}}

\begin{abstract}
Cerebral small vessel disease (CSVD) markers, specifically enlarged perivascular spaces (EPVS) and lacunae, present a unique challenge in medical image analysis due to their radiological mimicry. Standard segmentation networks struggle with feature interference and extreme class imbalance when handling these divergent targets simultaneously. To address these issues, we propose a morphology-decoupled framework where Zero-Initialized Gated Cross-Task Attention exploits dense EPVS context to guide sparse lacune detection. Furthermore, biological and topological consistency are enforced via a mixed-supervision strategy integrating Mutual Exclusion and Centerline Dice losses. Finally, we introduce an Anatomically-Informed Inference Calibration mechanism to dynamically suppress false positives based on tissue semantics. Extensive 5-folds cross-validation on the VALDO 2021 dataset ($N=40$) demonstrates state-of-the-art performance, notably surpassing task winners in lacunae detection precision ($71.1\%$, $p=0.01$) and F1-score ($62.6\%$, $p=0.03$). Furthermore, evaluation on the external EPAD cohort ($N=1762$) confirms the model's robustness for large-scale population studies. Code will be released upon acceptance.
\end{abstract}

\section{Introduction}

Cerebral small vessel disease (CSVD) is a primary cause of vascular dementia and a significant factor in stroke occurrence \cite{cannistraro_2019_cns, pantoni_2010_cerebral}. Among the neuroimaging manifestations of CSVD, enlarged perivascular spaces (EPVS) and lacunes of presumed vascular origin serve as critical biomarkers for the assessment of disease burden and progression \cite{wardlaw_2013_neuroimaging}. EPVS represent fluid-filled channels that reflect glymphatic dysfunction \cite{iliff_2012_a}, whereas lacunes indicate focal tissue infarction and permanent brain injury \cite{wardlaw_2013_neuroimaging}. Although these markers have distinct pathological origins, they frequently co-occur and exhibit significant radiological similarities. On all MRI sequences, both appear isointense to cerebrospinal fluid (CSF), manifesting as focal lesions with comparable signal intensities that complicate differentiation \cite{dubost_2019_3d}. Distinguishing the linear, tubular topology of EPVS from the ovoid morphology of lacunes is essential for accurate diagnosis, yet remains a challenge for automation \cite{wardlaw_2013_neuroimaging, sudre_2024_where}.

Current automated solutions predominantly quantify CSVD markers as isolated tasks. For instance, Dubost et al. \cite{dubost_2019_3d} pioneered weakly supervised networks specifically for EPVS, while top-performing methods in the VALDO challenge \cite{sudre_2024_where} often employ single-task segmentation architectures for lacunes. By ignoring the physiological correlation between these lesions, isolated models miss crucial contextual guidance. Extreme data imbalance, a recognized challenge for small CSVD markers \cite{sudre_2019_3d}, and the absence of biological priors further exacerbate these architectural flaws. While recent single-task methods have begun incorporating anatomical constraints \cite{li_2023_priorknowledgeinformed}, standard optimization on skewed distributions inherently biases joint models against sparse lacunes. Purely data-driven baselines routinely generate false positives in anatomically implausible regions. \cite{oktay_2018_anatomically}. 

To resolve these coupled challenges, we propose a morphology-decoupled multi-task framework. At its core, a Cross-Task Gated Attention module establishes a unidirectional information flow, leveraging the dense perivascular network as a spatial prior to guide sparse lacune detection without feature entanglement. Furthermore, we integrate explicit topological supervision and an adaptive distance-field calibration, effectively preserving structural integrity and suppressing clinically invalid predictions. Our contributions are as follows: 

\begin{enumerate}
    \item \textbf{Morphology-Decoupled Architecture:} We propose a shared-encoder, dual-decoder network featuring a Gated Cross-Task Attention mechanism. This design effectively disentangles tubular and spherical features, utilizing EPVS context to guide lacune detection while preventing feature collapse.
    \item \textbf{Mixed-Supervision \& Anatomy-Aware Constraints:} We introduce a hybrid optimization strategy that integrates fully supervised dense masks with weakly supervised regional counts for EPVS. Furthermore, it enforces biological consistency via Mutual Exclusion Loss and vascular topology through Centerline Dice supervision, mitigating the impact of label scarcity.
    \item \textbf{Anatomical Calibration:} We develop a Distance-Field Calibration module that modulates prediction confidence based on tissue semantics. This mechanism strictly suppresses false positives in exclusion zones.
    \end{enumerate}
We then evaluated performance on the VALDO dataset\cite{sudre_2024_where} and the generalisability of our proposed solution on a large multicentre external dataset \cite{ritchie_2019_the} in comparison with multiple state-of-the art algorithms.

\section{Methods}

\begin{figure}[htbp]
  \centering
  \includegraphics[width=\textwidth]{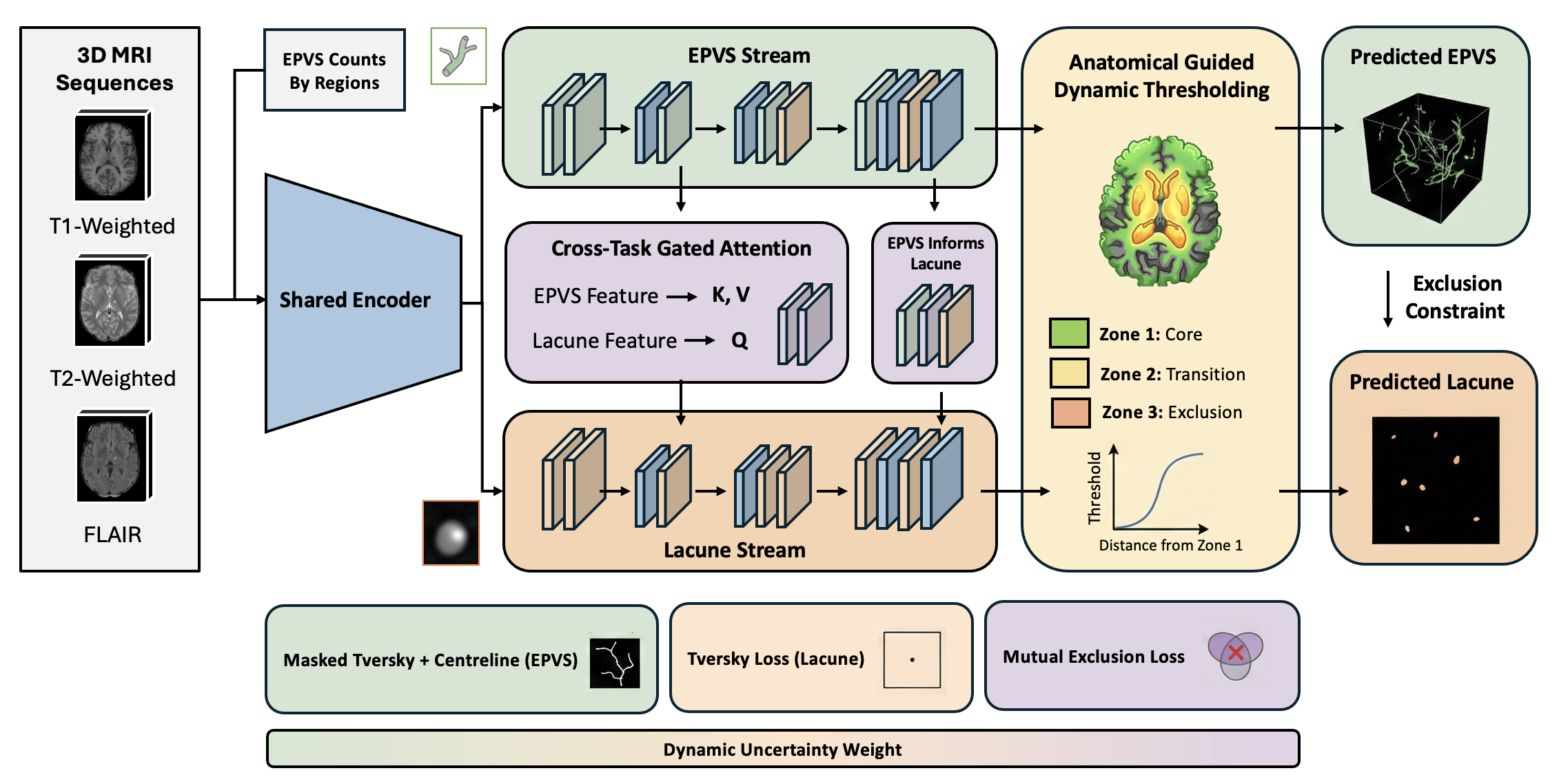}
  \caption{Overview of the Proposed Morphology-Decoupled Framework.}
  \label{fig:Network}
\end{figure}

\subsection{Multi-Task Network Architecture}

We present a multi-task 3D framework based on a Dynamic U-Net architecture (Fig. \ref{fig:Network}). To capture the distinct yet correlated morphological features of EPVS and lacunes, the network employs a shared encoder that branches into two task-specific decoders. This design enables the encoder to learn a common representation of the cerebral anatomy, while the decoders specialize in the fine-grained segmentation of tubular EPVS and ovoid lacunes, respectively\cite{moeskops_2016_deep}.

We hypothesize that regions exhibiting a high density of EPVS indicate severe local micro-vascular dysfunction, thereby acting as a spatial map that guides the network to locate the rarer, co-occurring lacunes. To model this dependency without the introduction of circular feedback, we introduce a Cross-Task Gated Attention module \cite{oktay_2018_attention}. This module enforces a unidirectional information flow from the EPVS stream to the Lacune stream at every up-sampling stage. By leveraging the local feature responses from the EPVS branch, the network explicitly guides the identification of sparse lacunar regions using the spatial context of perivascular spaces.

Let $F_{Lac}, F_{EPVS} \in \mathbb{R}^{C \times D \times H \times W}$ denote the Lacune (query) and EPVS (context) feature maps, respectively. We project these via $1\times1$ convolutions to obtain embeddings $Q$, $K$, and $V$. To minimize computational overhead, $Q$ and $K$ are mapped to a bottleneck dimension $C_{int} = C/r$ (with $r=4$), while $V$ retains the original channel dimension $C$. The spatial attention gate $G$ and the refined Lacune features $\hat{F}_{Lac}$ are computed as follows:

\begin{equation}
G = \sigma \left( \frac{1}{\sqrt{C_{int}}} \sum_{c=1}^{C_{int}} (Q_c \odot K_c) \right), \quad \hat{F}_{Lac} = F_{Lac} + G \odot V
\label{eq:attention_mechanism}
\end{equation}

\noindent where $\sigma$ is the sigmoid function and $\odot$ denotes element-wise multiplication. This mechanism aggregates global context across the bottleneck channels to highlight spatial regions where EPVS density structurally correlates with lacune formation.

This selective incorporation of EPVS context suppresses false positives in regions devoid of vascular pathology. Furthermore, we apply \textit{zero-initialization} to the value projection $V$ , ensuring the module initially acts as an identity mapping to facilitate stable task-specific convergence \cite{bachlechner_2020_rezero}. Finally, we employ deep supervision at three resolution levels to facilitate gradient flow \cite{zhou_2018_unet}.

\subsection{Anatomically-Informed Inference Calibration}

To mitigate false positives in anatomically implausible regions, we utilize FastSurfer \cite{henschel_2020_fastsurfer} to partition the brain volume into three reliability tiers: Zone 1 (Allowed), comprising white matter, deep grey matter and the brainstem; Zone 2 (Transition), covering the hippocampus and cerebellar white matter; and Zone 3 (Exclusion), encompassing the cortex, ventricles and extra-cerebral tissues.

We compute a one-sided truncated distance map $D(x)$, where $D(x)=0$ within Zone 1 and increases linearly with the Euclidean distance in exterior regions. To enforce anatomical consistency, we replace the standard static binarization threshold with a spatially adaptive decision boundary $T(x)$. For a given voxel $x$ with the network's predicted foreground probability $p(x)$, the final binary segmentation mask $M(x)$ is determined by the condition $M(x) = 1$ if $p(x) \geq T(x)$ and $0$ otherwise, where $T(x)$ is formulated as:

\begin{equation}
T(x) = \begin{cases} 
0.5 & \text{if } x \in \text{Zone 1} \\
0.5 + \lambda \cdot \tanh(\gamma \cdot D(x)) & \text{otherwise}
\end{cases}
\label{eq:dynamic_threshold}
\end{equation}

\noindent Here, the base value of $0.5$ represents the standard unbiased decision threshold for probabilistic predictions. The parameters $\lambda = 0.5$ and $\gamma = 0.5$ modulate the penalty magnitude and steepness, respectively. This formulation ensures that predictions in exclusion zones are heavily penalized, requiring significantly higher confidence (e.g., $p > 0.9$) to survive binarization, thereby effectively filtering cortical noise. Finally, we transition from voxel-level segmentation to object-level detection via Connected Component Analysis (CCA), extracting all candidate lesions with a minimum size of 1 voxel.

\subsection{Objective Function}

To address the extreme class imbalance, we employ the Tversky loss ($\mathcal{L}_{Tvk}$) with $\alpha=0.1$ and $\beta=0.9$ as the primary segmentation objective \cite{salehi_2017_tversky}, penalizing false negatives more heavily. For the EPVS branch, we explicitly incorporate the Soft-Centerline Dice loss ($\mathcal{L}_{clDice}$) \cite{suprosannashit_2021_cldice} to preserve the tubular topology of vascular structures. Additionally, a validity mask is applied to exclude unannotated regions from gradient computation. The task-specific losses, $\mathcal{L}_{Lac}$ and $\mathcal{L}_{EPVS}$, are computed by aggregating these objectives across all deep supervision scales.

To resolve radiological ambiguities between spatially proximate targets, we enforce a Mutual Exclusion loss ($\mathcal{L}_{excl}$). This term penalizes voxel-wise probabilistic overlap between the two classes:

\begin{equation}
\mathcal{L}_{excl} = \frac{1}{|\Omega|} \sum_{i \in \Omega} (p^{(i)}_{EPVS} \cdot p^{(i)}_{Lac})
\end{equation}

\noindent where $\Omega$ denotes the spatial domain and $p^{(i)}$ represents probability at voxel $i$.

Finally, we balance the multi-task learning using homoscedastic uncertainty weighting \cite{kendall_2018_multitask}. Let $s_t$ denote the learnable log-variance for task $t \in \{EPVS, Lac\}$. The total optimization objective is formulated as:

\begin{equation}
\mathcal{L}_{Total} = \sum_{t} (e^{-s_t} \mathcal{L}_t + s_t) + \lambda_{excl}\mathcal{L}_{excl}
\label{eq:total_loss}
\end{equation}

\noindent This formulation allows the network to dynamically down-weight tasks with high epistemic uncertainty during the early stages of training.

\section{Experiments and Results}

\subsection{Datasets}
We employed the VALDO 2021 challenge dataset ($N=40$) for training and voxel-level evaluation \cite{sudre_2024_where}. The dataset comprises co-registered T1, T2, and FLAIR sequences, resampled to an isotropic resolution of $1\,\text{mm}^3$. Annotations include fully segmented Lacune masks and a mixture of dense masks ($N=12$) and weak regional counts ($N=28$) for EPVS. To assess clinical generalizability, we further utilized the external EPAD cohort ($N=1762$) \cite{ritchie_2019_the}. Given the absence of voxel-wise ground truth, evaluation on EPAD relies on weak labels, patient level lacune presence and EPVS visual rating \cite{potter_2015_cerebral}.

\subsection{Implementation Details}
The framework was implemented in PyTorch utilizing the MONAI library \cite{cardoso_2022_monai}, and trained on an NVIDIA A100 (80GB) GPU. Images underwent z-score intensity normalization. To address volumetric imbalance, we employed a dual-stream sampling strategy \cite{isensee_2020_nnunet}, explicitly balancing Lacune- and EPVS-centered crops to training patches of size $96^3$. Data augmentation included random spatial flips and gamma contrast adjustments. Optimization was performed using AdamW (learning rate $1\times10^{-4}$, weight decay $1\times10^{-5}$) for 200 epochs. Inference utilized a sliding window approach with an ROI size of $128^3$ and overlap 0.6, evaluated via 5-fold cross-validation.

\subsection{Evaluation Metrics}
Similarly to the VALDO evaluation, for instance-level detection, a Lacune is classified as a True Positive if its centroid falls within $5\,\text{mm}$ of the ground truth, whereas EPVS detection requires an Intersection over Union (IoU) $> 10\%$. We report Precision, Recall, F1-score, and False Positives per subject. Segmentation quality over true positive elements was assessed with the Dice Similarity Coefficient (DSC) and Normalized Surface Distance (NSD, tolerance $1\,\text{mm}$). For the EPAD cohort, we evaluate Lacune clinical utility using Balanced Accuracy (presence detection), Mean Absolute Error (MAE, count disparity), and Pearson's Correlation ($r$) for global burden. For EPVS, we compute Spearman's Rank Correlation ($\rho$) across the basal ganglia (BG), centrum semi-ovale (CSO), and mid-brain (MB) to account for the categorical Potter scale. Furthermore, confidence intervals (CIs) for both tasks were calculated using bootstrapping with 2000 iterations. Statistical significance of performance differences was determined via a paired Wilcoxon signed-rank test ($p < 0.05$).

\subsection{Main Results}

\begin{table}[t]
\caption{Comparison with SOTA methods on the VALDO and EPAD datasets. \textbf{Bold} indicates best numerical results. * denotes statistical significance. Spearman's rank correlation and Pearson's correlation coefficient are denoted by $\rho$ and $r$, respectively.}
\label{tab:main_results}
\centering
\resizebox{\textwidth}{!}{%
\begin{tabular}{l cc cccc ccc}
\toprule
\multicolumn{1}{c}{\multirow{2}{*}{\textbf{Task/Model}}} & \multicolumn{6}{c}{\textbf{VALDO}} & \multicolumn{3}{c}{\textbf{EPAD}} \\
\cmidrule(lr){2-7} \cmidrule(lr){8-10}
& \multicolumn{2}{c}{\textbf{Segmentation}} & \multicolumn{4}{c}{\textbf{Detection}} & \multicolumn{3}{c}{} \\
\cmidrule(lr){2-3} \cmidrule(lr){4-7}
\textit{Task 1: EPVS} & DSC (\%) $\uparrow$ & NSD (\%)$\uparrow$ & Recall (\%)$\uparrow$ & Precision (\%)$\uparrow$ & F1 (\%)$\uparrow$ & FP/Subject $\downarrow$ & $\rho$ (BG) $\uparrow$ & $\rho$ (CSO) $\uparrow$ & $\rho$ (MB) $\uparrow$ \\
\midrule
DYN Unet & $36.9 \pm 9.7$ & $56.9 \pm 12.5$ & $50.2 \pm 1.8$ & $54.9 \pm 16.2$ & $48.6 \pm 9.8$ & $21.0 \pm 11.9$ & $0.13 \pm 0.02$ & $0.17 \pm 0.02$ & $0.10 \pm 0.02$ \\
MedNeXt & $37.9 \pm 10.2$ & $55.6 \pm 12.0$ & $52.4 \pm 7.5$ & $46.1 \pm 13.2$ & $45.7 \pm 8.7$ & $30.6 \pm 19.5$ & $0.09 \pm 0.02$ & $0.09 \pm 0.02$ & $0.06 \pm 0.02$ \\
Swin-UNETR V2 & $41.9 \pm 9.2$ & $62.2 \pm 11.9$ & $\mathbf{58.4 \pm 5.6}$ & $55.0 \pm 19.7$ & $50.5 \pm 10.7$ & $30.4 \pm 23.6$ & $0.02 \pm 0.02$ & $0.05 \pm 0.02$ & $0.03 \pm 0.02$ \\
VISTA-3D & $36.3 \pm 8.0$ & $57.6 \pm 9.8$ & $51.0 \pm 8.3$ & $49.7 \pm 10.7$ & $47.7 \pm 9.1$ & $23.7 \pm 7.6$ & $0.14 \pm 0.03$ & $0.17 \pm 0.02$ & $0.10 \pm 0.03$ \\
VALDO Winner & $\mathbf{42.8 \pm 9.9}$ & $\mathbf{63.7 \pm 11.5}$ & $53.2 \pm 4.4$ & $58.3 \pm 13.5$ & $50.5 \pm 9.2$ & $17.9 \pm 10.0$ & $0.10 \pm 0.03$ & $0.13 \pm 0.03$ & $0.05 \pm 0.02$ \\
Our Method & $38.1 \pm 6.5$ & $56.7 \pm 8.4$ & $49.8 \pm 10.0$ & $\mathbf{67.4 \pm 17.2}$ & $\mathbf{53.7 \pm 9.6}$ & $\mathbf{15.6 \pm 10.8}$ & $\mathbf{0.22 \pm 0.02}$ & $\mathbf{0.29 \pm 0.03}$ & $\mathbf{0.11 \pm 0.02}$ \\
\midrule
\textit{Task 3: Lacune} & DSC (\%) $\uparrow$ & NSD (\%) $\uparrow$ & Recall (\%) $\uparrow$ & Precision (\%) $\uparrow$ & F1 (\%) $\uparrow$ & FP/Subject $\downarrow$ & BAcc (\%) $\uparrow$ & MAE $\downarrow$ & $r$ (Glob)$\uparrow$ \\
\midrule
DYN Unet & $27.7 \pm 7.6$ & $39.1 \pm 9.2$ & $55.8 \pm 14.3$ & $34.6 \pm 18.5$ & $35.7 \pm 11.7$ & $4.3 \pm 2.7$ & $51.6 \pm 0.9$ & $2.6 \pm 0.04$ & $0.11 \pm 0.03$ \\
MedNeXt & $30.4 \pm 4.9$ & $43.0 \pm 5.1$ & $\mathbf{66.2 \pm 13.3}$ & $30.3 \pm 9.7$ & $34.5 \pm 5.5$ & $6.1 \pm 2.5$ & $52.8 \pm 1.0$ & $0.5 \pm 0.01$ & $0.11 \pm 0.07$ \\
Swin-UNETR V2 & $31.3 \pm 7.1$ & $44.2 \pm 8.3$ & $59.1 \pm 9.4$ & $30.5 \pm 12.2$ & $35.3 \pm 9.5$ & $5.1 \pm 1.8$ & $48.7 \pm 2.1$ & $2.0 \pm 0.10$ & $-0.03 \pm 0.01$ \\
VISTA-3D & $29.1 \pm 6.4$ & $40.2 \pm 8.4$ & $64.5 \pm 20.4$ & $34.5 \pm 10.1$ & $40.1 \pm 9.6$ & $4.0 \pm 1.3$ & $51.7 \pm 2.1$ & $0.8 \pm 0.04$ & $0.06 \pm 0.04$ \\
VALDO Winner & $38.3 \pm 4.7$ & $45.9 \pm 7.5$ & $59.5 \pm 18.9$ & $37.8 \pm 13.5$ & $42.2 \pm 13.4$ & $2.2 \pm 0.9$ & $52.1 \pm 1.1$ & $2.6 \pm 0.04$ & $0.06 \pm 0.04$ \\
Our Method & $\mathbf{42.4 \pm 11.3}$ & $\mathbf{58.5 \pm 15.9}$ & $58.4 \pm 18.1$ & $\mathbf{71.1 \pm 17.3^*}$ & $\mathbf{62.6 \pm 17.1^*}$ & $\mathbf{0.7 \pm 0.9}$ & $\mathbf{64.9 \pm 2.1}$ & $\mathbf{0.2 \pm 0.01}$ & $\mathbf{0.24 \pm 0.05}$ \\
\bottomrule
\end{tabular}%
}
\end{table}

Table \ref{tab:main_results} benchmarks our framework against strong MONAI baselines (Such as Swin-UNETR \cite{hatamizadeh_2022_swin} and VISTA-3D \cite{he_2024_vista3d}) and our re-implementations of the task-specific VALDO 2021 challenge winners \cite{sudre_2024_where}.  For the baseline methods providing only segmentation, probabilistic outputs were thresholded and connected components extracted. For EPVS detection, we achieve an F1-score of $53.7 \pm 9.6\%$, numerically exceeding the challenge winner($50.5 \pm 9.2\%$) despite a lower recall ($49.8 \pm 10.0\%$ compared to $53.2 \pm 4.4\%$). In Lacune detection, our approach demonstrates a statistically significant improvement in both Precision ($71.1 \pm 17.3\%$, $p = 0.01$) and F1-score ($62.6 \pm 17.1\%$, $p = 0.03$) versus the challenge winner. External evaluation on the EPAD cohort further validates clinical robustness. For the Lacune task, our model achieves a state-of-the-art Balanced Accuracy of $64.9 \pm 2.1\%$, a global correlation of $r=0.24 \pm 0.05$, and a minimal MAE of $0.2 \pm 0.01$. Regarding EPVS burden estimation, our method achieves the highest regional concordance with visual ratings (e.g., CSO $\rho=0.29 \pm 0.03$), substantially outperforming best baselines such as VISTA-3D (CSO $\rho=0.17 \pm 0.02$).

\begin{figure}[htbp]
  \centering
  \includegraphics[width=\textwidth]{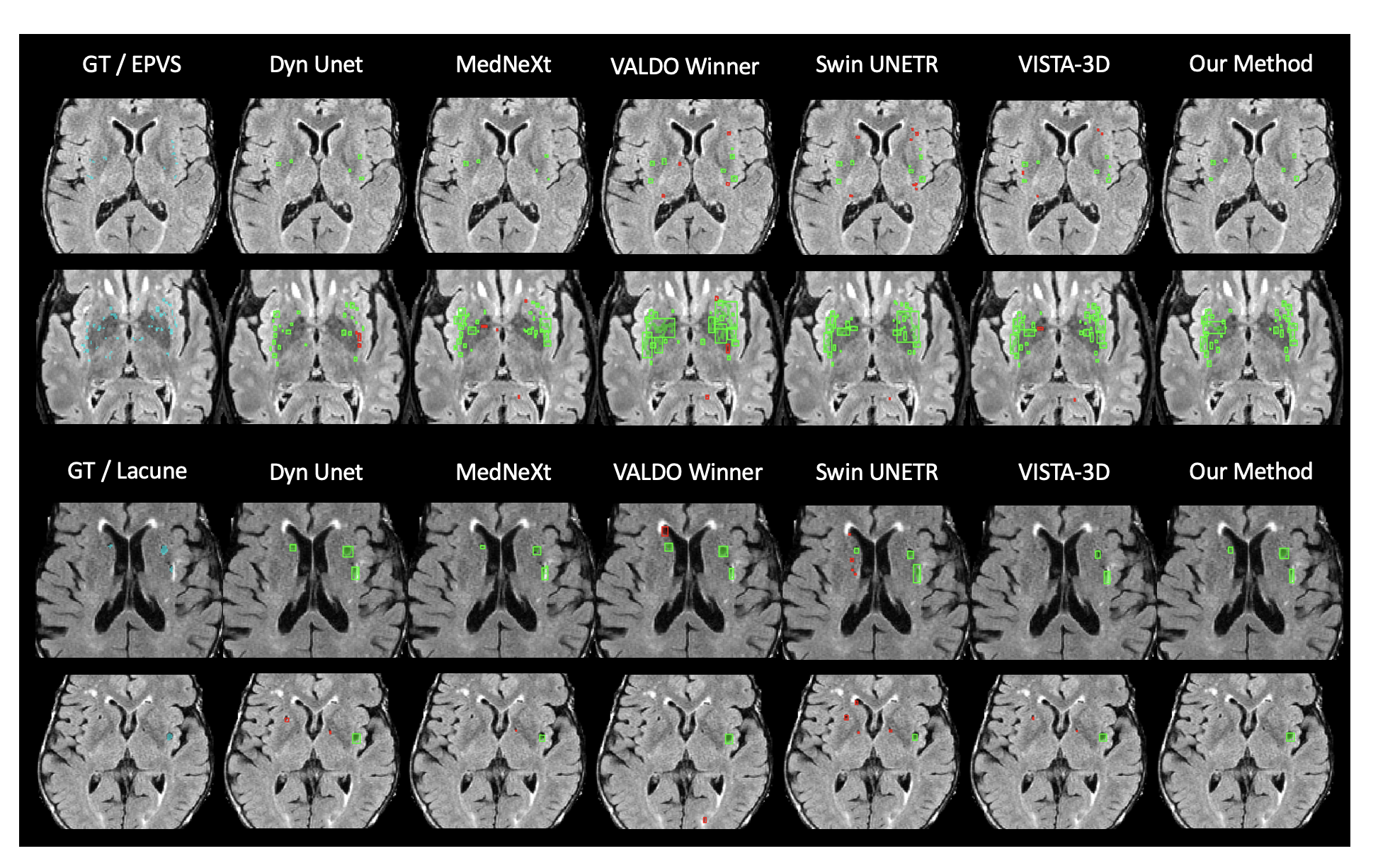}
  \caption{Qualitative visualization on the VALDO dataset. Ground truth lesions (cyan), true positive predictions (green), and false positive predictions (red) are shown.}
  \label{fig:qualitative}
\end{figure}

Fig. \ref{fig:qualitative} visually corroborates these metrics. While competing methods frequently generate false positives in cortical regions, our anatomically calibrated approach demonstrates superior specificity. It accurately localizes sparse lacunes and dense EPVS with minimal background noise, filtering out radiological mimics where baselines fail. However, this stringent calibration entails a trade-off with sensitivity, as it occasionally misses small or faint lesions.

\subsection{Ablation Studies}

\begin{table}[t]
\caption{Ablation of objective functions on VALDO. \textbf{Bold} indicates numercial best}
\label{tab:ablation_loss}
\centering
\resizebox{\textwidth}{!}{%
\begin{tabular}{ccc cc cccc}
\toprule
\multicolumn{3}{c}{\textbf{Task/model}} & \multicolumn{2}{c}{\textbf{Segmentation}} & \multicolumn{4}{c}{\textbf{Detection}} \\
\cmidrule(lr){1-3} \cmidrule(lr){4-5} \cmidrule(lr){6-9}
$\mathbf{L_{MT}}$ & $\mathbf{L_{exc}}$ & $\mathbf{L_{clDice}}$ & DSC (\%) $\uparrow$ & NSD (\%)$\uparrow$ & Recall (\%)$\uparrow$ & Precision (\%)$\uparrow$ & F1 (\%)$\uparrow$ & FP/Subject $\downarrow$ \\
\midrule
\multicolumn{9}{l}{\textit{Task 1: EPVS}} \\
\midrule
\checkmark & & & $35.6 \pm 9.4$ & $54.5 \pm 12.0$ & $45.6 \pm 7.6$ & $58.9 \pm 17.3$ & $47.7 \pm 9.6$ & $17.1 \pm 14.2$ \\
\checkmark & \checkmark & & $35.7 \pm 9.5$ & $54.2 \pm 11.9$ & $46.9 \pm 7.7$ & $59.2 \pm 17.5$ & $48.9 \pm 9.8$ & $15.3 \pm 13.6$ \\
\checkmark & & \checkmark & $\mathbf{36.9 \pm 9.7}$ & $\mathbf{54.7 \pm 12.1}$ & $46.6 \pm 7.7$ & $\mathbf{62.6 \pm 19.1}$ & $49.2 \pm 9.9$ & $\mathbf{16.4 \pm 12.3}$ \\
\checkmark & \checkmark & \checkmark & $34.3 \pm 8.0$ & $51.1 \pm 12.4$ & $\mathbf{47.1 \pm 7.5}$ & $60.4 \pm 10.7$ & $\mathbf{50.0 \pm 6.2}$ & $16.5 \pm 11.6$ \\
\midrule
\multicolumn{9}{l}{\textit{Task 3: Lacune}} \\
\midrule
\checkmark & & & $\mathbf{45.3 \pm 15.5}$ & $46.4 \pm 17.3$ & $50.0 \pm 12.8$ & $46.3 \pm 22.0$ & $45.8 \pm 14.9$ & $3.9 \pm 2.7$ \\
\checkmark & \checkmark & & $36.3 \pm 13.9$ & $\mathbf{51.4 \pm 19.2}$ & $\mathbf{68.1 \pm 17.5}$ & $50.1 \pm 23.8$ & $52.7 \pm 17.2$ & $2.2 \pm 1.5$ \\
\checkmark & & \checkmark & $35.0 \pm 13.4$ & $45.2 \pm 16.8$ & $58.4 \pm 15.0$ & $49.1 \pm 23.3$ & $47.0 \pm 15.3$ & $3.3 \pm 2.3$ \\
\checkmark & \checkmark & \checkmark & $36.5 \pm 14.0$ & $\mathbf{51.4 \pm 19.2}$ & $62.7 \pm 16.1$ & $\mathbf{57.8 \pm 27.5}$ & $\mathbf{56.1 \pm 18.3}$ & $\mathbf{1.0 \pm 0.7}$ \\
\bottomrule
\end{tabular}%
}
\end{table}

Table \ref{tab:ablation_loss} evaluates the proposed objective functions. The baseline model, trained solely with standard segmentation losses ($L_{MT}$), exhibits limited sensitivity for Lacunes (Recall: $50.0 \pm 12.8\%$). Incorporating the Mutual Exclusion Loss ($L_{exc}$) explicitly addresses the biological inconsistency of overlapping labels. This constraint yields an improvement in Lacune Recall, rising to $68.1 \pm 17.5\%$, while simultaneously reducing False Positives (FPs) from $3.9$ to $2.2$ per subject. Conversely, the Soft-Centerline Dice loss ($L_{clDice}$) specifically benefits the tubular EPVS structures (Task 1), boosting Precision to its peak of $62.6 \pm 19.1\%$. The combined objectives provides the most robust trade-off, achieving the lowest FP rate ($1.0 \pm 0.7$) for Lacunes while maintaining optimal F1-scores across tasks.

Table \ref{tab:ablation_arch} validates the necessity of our unified framework design. Standard fully supervised Single-Task Learning (STL-Full) fails completely on the imbalanced EPVS class, resulting in a prohibitive FP rate of $94.8 \pm 91.9$ and an F1-score of only $18.3 \pm 14.8\%$. The introduction of mixed supervision (STL-Mix) effectively mitigates this, recovering the F1-score to $48.6 \pm 9.8\%$. Furthermore, transitioning from a shared-decoder (MTL-Shared) to an attention-decoupled architecture (MTL-Gated-Attn) resolves feature interference between the divergent targets. This gated-attention split yields consistent gains, improving the F1-score for EPVS ($50.0\% \to 51.8\%$) and Lacunae ($56.1\% \to 58.2\%$) by allowing independent optimization of tubular and round/ovoid representations.

The integration of Anatomically-Informed Inference Calibration refines further the output (Table \ref{tab:ablation_arch}, bottom row). by filtering out predictionsin biologically implausible regions. For Lacunes, it results in a reduction of the FP rate to $0.7 \pm 0.9$ with minimal changes in recall confirming the validity of the anatomical guidance. 

\begin{table}[t]
\caption{Comparison of supervision strategies, cross-task attention decoupling, and anatomical calibration on the VALDO dataset. \textbf{Bold} indicates numerical best results.}
\label{tab:ablation_arch}
\centering
\resizebox{\textwidth}{!}{%
\begin{tabular}{l c c cc cccc}
\toprule
\multicolumn{3}{c}{\textbf{Task/model}} & \multicolumn{2}{c}{\textbf{Segmentation}} & \multicolumn{4}{c}{\textbf{Detection}} \\
\cmidrule(lr){1-3} \cmidrule(lr){4-5} \cmidrule(lr){6-9}
Method & Decoder & Anatomical & DSC (\%)$\uparrow$ & NSD (\%)$\uparrow$ & Recall (\%)$\uparrow$ & Precision (\%)$\uparrow$ & F1 (\%)$\uparrow$ & FP/Subject $\downarrow$ \\
\midrule
\multicolumn{9}{l}{\textit{Task 1: EPVS}} \\
\midrule
STL-Full & -- & -- & $12.4 \pm 17.3$ & $20.4 \pm 27.5$ & $25.9 \pm 17.9$ & $21.4 \pm 18.0$ & $18.3 \pm 14.8$ & $94.8 \pm 91.9$ \\
STL-Mix & -- & -- & $36.9 \pm 9.7$ & $56.9 \pm 12.5$ & $\mathbf{50.2 \pm 1.8}$ & $54.9 \pm 16.2$ & $48.6 \pm 9.8$ & $21.0 \pm 11.9$ \\
MTL & Shared & & $34.3 \pm 8.0$ & $51.1 \pm 12.4$ & $47.1 \pm 8.5$ & $60.4 \pm 10.7$ & $50.0 \pm 6.2$ & $16.5 \pm 11.6$ \\
MTL & Shared & \checkmark & $35.3 \pm 7.1$ & $52.8 \pm 10.8$ & $47.1 \pm 8.5$ & $60.8 \pm 12.7$ & $49.8 \pm 6.7$ & $16.1 \pm 11.1$ \\
MTL & Gated-Attn & & $37.5 \pm 7.1$ & $56.3 \pm 9.0$ & $50.0 \pm 10.3$ & $63.5 \pm 15.3$ & $51.8 \pm 9.7$ & $15.9 \pm 13.1$ \\
MTL & Gated-Attn & \checkmark & $\mathbf{38.1 \pm 6.5}$ & $\mathbf{56.7 \pm 8.4}$ & $49.8 \pm 10.0$ & $\mathbf{67.4 \pm 15.2}$ & $\mathbf{53.7 \pm 9.6}$ & $\mathbf{15.6 \pm 10.8}$ \\
\midrule
\multicolumn{9}{l}{\textit{Task 3: Lacunae}} \\
\midrule
STL & -- & -- & $27.7 \pm 7.6$ & $39.1 \pm 9.2$ & $55.8 \pm 14.3$ & $34.6 \pm 18.5$ & $35.7 \pm 11.7$ & $4.3 \pm 2.7$ \\
MTL & Shared & & $36.5 \pm 14.0$ & $51.4 \pm 19.2$ & $\mathbf{62.7 \pm 16.1}$ & $57.8 \pm 27.5$ & $56.1 \pm 18.3$ & $1.0 \pm 0.7$ \\
MTL & Shared & \checkmark & $35.6 \pm 14.5$ & $49.8 \pm 20.5$ & $\mathbf{62.7 \pm 16.1}$ & $57.4 \pm 27.4$ & $55.9 \pm 18.3$ & $1.0 \pm 0.6$ \\
MTL & Gated-Attn & & $\mathbf{43.3 \pm 11.6}$ & $57.3 \pm 16.0$ & $59.0 \pm 18.3$ & $65.6 \pm 19.2$ & $58.2 \pm 17.9$ & $1.2 \pm 0.9$ \\
MTL & Gated-Attn & \checkmark & $42.4 \pm 11.3$ & $\mathbf{58.5 \pm 15.9}$ & $58.4 \pm 18.1$ & $\mathbf{71.1 \pm 17.3}$ & $\mathbf{62.6 \pm 17.1}$ & $\mathbf{0.7 \pm 0.9}$ \\
\bottomrule
\end{tabular}%
}
\end{table}

\section{Conclusion}

In this paper, we presented a morphology-informed framework for the joint detection of lacunes and EPVS, addressing the dual challenges of extreme data imbalance and radiological mimicry. By applying mixed supervision with anatomically-informed inference calibration, our approach effectively enforces biological consistency while preserving the distinct topological features of tubular EPVS and round/ovoid lacunae. Quantitative benchmarking on the VALDO dataset demonstrates state-of-the-art overall performance, surpassing challenge winners in precision and F1-score. Moreover, successful deployment on a large multi-center EPAD cohort validates the model's robustness for population-level studies, demonstrating its capacity to extract reliable regional and global metrics across diverse demographics. This work provides an automated, highly reliable tool for quantifying vascular burden, paving the way for broad epidemiological research on cerebrovascular progression.

%
%
%
\bibliographystyle{splncs04}
\bibliography{MICCAI}

\end{document}